\pdfoutput=1
\documentclass{article}

\PassOptionsToPackage{numbers, compress}{natbib}
\usepackage[preprint]{nips_2018}
\usepackage{multirow}
\usepackage{amsmath,amssymb}
\usepackage{wrapfig}
\usepackage{wrapfig,lipsum,booktabs}
\usepackage{soul}
\usepackage{wrapfig,booktabs}
\usepackage{color}
\usepackage{algorithm}
\usepackage{algorithmic}

\usepackage[utf8]{inputenc} 
\usepackage[T1]{fontenc}    
\usepackage{hyperref}       
\usepackage{url}            
\usepackage{booktabs}       
\usepackage{amsfonts}       
\usepackage{nicefrac}       
\usepackage{microtype}      
\usepackage{graphicx}

\title{DADA: Deep Adversarial Data Augmentation for Extremely Low Data Regime Classification}

\author{
  Xiaofeng Zhang\\
  University of Science and Technology of China\\
  Hefei, Anhui, China\\
  \texttt{zxfeng@mail.ustc.edu.cn}\\
  \And
  Zhangyang Wang\\
  Texas A\&M University\\
  College Station, TA, USA\\
  \texttt{atlaswang@tamu.edu}\\
  \And
  Dong Liu\\
  University of Science and Technology of China\\
  Hefei, Anhui, China\\
  \texttt{dongeliu@ustc.edu.cn}\\
  \And
  Qing Ling\\
  Sun Yat-Sen University\\
  Guangzhou, Guangdong, China\\
  \texttt{lingqing556@mail.sysu.edu.cn}\\
}

\begin{document}

\maketitle

\begin{abstract}
  Deep learning has revolutionized the performance of classification, but meanwhile demands sufficient labeled data for training. Given insufficient data, while many techniques have been developed to help combat overfitting, the challenge remains if one tries to train deep networks, especially in the ill-posed \textit{extremely low data regimes}: only a small set of labeled data are available, and nothing -- including unlabeled data -- else. Such regimes arise from practical situations where not only data labeling but also data collection itself is expensive. We propose a deep adversarial data augmentation (DADA) technique to address the problem, in which we elaborately formulate data augmentation as a problem of training a class-conditional and supervised generative adversarial network (GAN). Specifically, a new discriminator loss is proposed to fit the goal of data augmentation, through which both real and augmented samples are enforced to contribute to and be consistent in finding the decision boundaries. Tailored training techniques are developed accordingly. To quantitatively validate its effectiveness, we first perform extensive simulations to show that DADA substantially outperforms both traditional data augmentation and a few GAN-based options. We then extend experiments to three real-world small labeled datasets where existing data augmentation and/or transfer learning strategies are either less effective or infeasible. All results endorse the superior capability of DADA in enhancing the generalization ability of deep networks trained in practical extremely low data regimes. Source code is available at \url{https://github.com/SchafferZhang/DADA}.
\end{abstract}

\section{Introduction}
The performance of classification and recognition has been tremendously revolutionized by the prosperity of deep learning \cite{krizhevsky2012imagenet}. 
Deep learning-based classifiers can reach unprecedented accuracy given that there are sufficient labeled data for training. Meanwhile, such a blessing can turn into a curse: in many realistic settings where either massively annotating labels is a labor-intensive task, or only limited datasets are available, a deep learning model will easily overfit and generalizes poorly. Many techniques have been developed to help combat overfitting with insufficient data, ranging from classical data augmentation \cite{perez2017effectiveness}, to dropout \cite{krizhevsky2012imagenet} and other structural regularizations \cite{han2015learning}, to pre-training \cite{erhan2010does}, transfer learning \cite{raina2007self} and semi-supervised learning \cite{kingma2014semi}. However in low data regimes, even these techniques will fall short, and the resulting models usually cannot capture all possible input data variances and distinguish them from nuisance variances. The high-variance gradients also cause popular training algorithms, e.g., stochastic gradient descent, to be extremely unstable.

In this paper, we place ourself in front of an even more ill-posed and challenging problem: \textit{how to learn a deep network classifier, where the labeled training set is high-dimensional but small in sample size?} Most existing methods in the low data regimes deal with the scarcity of labeled data; however, they often assume the help from abundant unlabeled samples in the same set, or (labeled or unlabeled) samples from other similar datasets, enabling various \textit{semi-supervised} or \textit{transfer learning} solutions. Different from them, we investigate a less-explored and much more daunting task setting of \textbf{extremely low data regimes}: besides the given small amount of labeled samples, neither unlabeled data from the same distribution nor data from similar distributions are assumed to be available throughout training. In other words, \textbf{we aim to train a deep network classifier from scratch, using only the given small number of labeled data and nothing else}. Our only hope lies in maximizing the usage of the given small training set, by finding nontrivial and semantically meaningful re-composition of sample information that helps us characterize the underlying distribution. 
The extremely low data regimes for classification are ubiquitous in and have blocked many practical or scientific fields, where \textbf{not only data labeling, but data collection itself is also expensive to scale up}. For example, image subjects from military and medical imagery are usually expensive to collect, and often admit quite different distributions from easily accessible natural images. While we mostly focus on image classification/visual recognition in this paper, our methodology can be readily extended to classifying non-image data in extremely low data regimes; we will intentionally present one such example of electroencephalographic (EEG) signal classification in Section \ref{sec_exp}.
To resolve the challenges, we have made multi-fold technical contributions in this paper:
\begin{itemize}
\item For learning deep classifiers in extremely low data regimes, we focus on boosting the effectiveness of data augmentation, and introduce learning-based data augmentation, that can be optimized for classifying general data without relying on any domain-specific prior or unlabeled data. The data augmentation module and the classifier are formulated and learned together as a fully-supervised generative adversarial network (GAN). We call the proposed framework \textit{Deep Adversarial Data Augmentation} (\textbf{DADA}). 
\item We propose a new loss function for the GAN discriminator, that not only learns to classify real images, but also enforces fine-grained classification over multiple ``fake classes''. That is referred to as the \textbf{2$k$ loss}, in contrast to the \textit{$k$+1 loss} used by several existing GANs (to be compared in the context later). The novel loss function is motivated by our need of data augmentation: \textit{the generated augmented (``fake'') samples need to be discriminative among classes too, and the decision boundaries learned on augmented samples shall align consistently with those learned on real samples}. We show in experiments that the 2$k$ loss is critical to boost the overall classification performance. 
\item We conduct extensive simulations on CIFAR-10, CIFAR-100, and SVNH, to train deep classifiers in the extremely low data regimes, demonstrating significant performance improvements through DADA compared to using traditional data augmentation. To further validate the practical effectiveness of DADA, we train deep classifiers on three real-world small datasets: the Karolinska Directed Emotional Faces (KDEF) dataset for the facial expression recognition task, a Brain-Computer Interface (BCI) Competition dataset for the EEG brain signal classification task, and the Curated Breast Imaging Subset of the Digital Database for Screening Mammography (CBIS-DDSM) dataset for the tumor classification task. For all of them, DADA leads to highly competitive generalization performance. 
\end{itemize}
\section{Related Work}
\label{gen_inst}
\vspace{-0.5em}
\subsection{Generative Adversarial Networks}
Generative Adversarial Networks (GANs) \cite{goodfellow2014generative} have gathered a significant amount of attention due to their ability to learn generative models of multiple natural image datasets. The original GAN model and its many successors are unsupervised: their discriminators have a single probabilistic realness output attempting to decipher whether an input image is real or generated (a.k.a. fake). Conditional GAN \cite{mirza2014conditional} generates data conditioned on class labels via label embeddings in both discriminator and generator. Conditioning generated samples on labels sheds light the option of semi-supervised classification using GANs. In \cite{odena2016semi}, the semi-supervised GAN has the discriminator network to output class labels, leading to a $k+1$ class loss function consisting of $k$ class labels if the sample is decided to be real, and a single extra class if the sample is decided to be fake. Such a structured $k+1$ loss has been re-emphasized in \cite{salimans2016improved} to provide more informed training that leads to generated samples capturing class-specific variances better. 
Even with the proven success of GANs for producing realistic-looking images, tailoring GANs for classification is not as straightforward as it looks like \cite{dai2017good}. The first question would naturally be: is GAN really creating semantically novel compositions, or simply memorizing its training samples (or adding trivial nuisances)? Fortunately, there seems to be empirical evidence that GANs perform at least some non-trivial modeling of the unknown distribution and are able to interpolate in the latent space \cite{radford2015unsupervised,arora2017gans}. However, previous examinations also reported that the diversity of generated samples is far poorer than the true training dataset. In \cite{santurkar2017classification}, the authors tried several unconditional GANs to synthesize samples, on which they trained image classifiers. They reported that the accuracies achieved by such classifiers were comparable only to the accuracy of a classifier trained on a 100 (or more) subsampled version of the true dataset, and the gap cannot be reduced by drawing more samples from the GANs. Despite many insights revealed, the authors did not consider low data regimes. More importantly, they focused on a different goal on using classification performance to measure the diversity of generated data. As a result, they neither considered class-conditional GANs, nor customized any GAN structure for the goal of classification-driven data augmentation. Besides, GANs also have hardly been focused towards non-image subjects.
\vspace{-1em}
\subsection{Deep Learning on Small Samples}
Training a deep learning classifier on small datasets is a topic of wide interests in the fields of fine-grained visual recognition \cite{lin2015bilinear}, few-shot learning \cite{mehrotra2017generative}, and life-long learning in new environments \cite{shu2017lifelong}. Hereby we review and categorize several mainstream approaches. 
\vspace{0.2em}

\noindent \textit{Dimensionality Reduction and Feature Selection.} A traditional solution to overfitting caused by high dimension, low sample size data is to perform dimensionality reduction or feature selection as pre-processing, and to train (deep) models on the new feature space. Such pre-processing has become less popular in deep learning because the latter often emphasizes end-to-end trainable pipelines. A recent work \cite{liu2017deep} performed the joint training of greedy feature selection and a deep classifier; but their model was designed for bioinformatics data (attributed vectors) and it was unclear how a similar model can be applied to raw images. 
\vspace{0.2em}

\noindent \textit{Pre-training and Semi-Supervised Learning.} Both pre-training and semi-supervised learning focus on improving classification with smalled labeled samples, by utilizing extra data from the same training distribution but is unlabeled. Greedy pre-training with larger unlabeled data, e.g., via auto-encoders, could help learn (unsupervised) feature extractors and converge to a better generalizing minimum \cite{erhan2010does}. In practice, pre-training is often accompanied with data augmentation \cite{vincent2010stacked}. Semi-supervised learning also utilizes extra unlabeled data, while unlabeled data contribute to depicting data density and thus locating decision boundaries within low-density regions; see \cite{lee2013pseudo,kingma2014semi,rasmus2015semi,salimans2016improved}. However, note that both pre-training and semi-supervised learning rely heavily on the abundance of unlabeled data: they are motivated by the same hypothesis that while labeling data is difficult, collecting unlabeled data remains to be a cheap task. While the hypothesis is valid in many computer vision tasks, it may not always stand true and differs from our target -- extremely low data regimes.
\vspace{0.2em}

\noindent \textit{Transfer Learning.} Compared to the above two, transfer learning admits a more relaxed setting: using unlabeled data from a similar or overlapped distribution (a.k.a. source domain), rather than from the same target distribution as labeled samples (a.k.a. target domain). For standard visual recognition, common visual patterns like edges are often shared between different natural image datasets. This makes a knowledge transfer between such datasets promising \cite{raina2007self}, even though their semantics are not strictly tied. Empirical study \cite{wagner2013learning} showed that, the weight transfer from deep networks trained on a source domain with abundant (labeled) data can boost visual recognition on a target domain where labeled samples are scarce. It is, however, unclear whether transfer or how much learning will help, if the source and target domains possess notable discrepancy. 
\vspace{0.2em}

\noindent\textit{Data Augmentation.} Data augmentation is an alternative strategy to bypass the unavailability of labeled training data, by artificially synthesizing new labeled samples from existing ones. Traditional data augmentation techniques rely on a very limited set of known invariances that are easy to invoke, and adopt ad-hoc, minor perturbations that will not change labels. For instance, in the case of image classification, typical augmentations include image rotation, lighting/color tone modifications, rescaling, cropping, or as simple as adding random noise \cite{krizhevsky2012imagenet}. However, such empirical label-preserving transformations are often unavailable in non-image domains. A latest work \citep{ratner2017learning} presented a novel direction to select and compose pre-specified base data transformations (such as rotations, shears, central swirls for images) into a more sophisticated ``tool chain'' for data augmentation, using generative adversarial training. They achieve highly promising results on both image and text datasets, but need the aid of unlabeled data in training (the same setting as in \cite{salimans2016improved}). \textit{We experimentally compare the method \citep{ratner2017learning} and DADA and analyze their more differences in Section 5.3.}


Few efforts went beyond encoding priori known invariances to explore more sophisticated, learning-based augmentation strategies. Several semi-supervised GANs, e.g., \cite{salimans2016improved}, could also be viewed as augmented unlabeled samples from labeled ones. A Bayesian Monte Carlo algorithm for data augmentation was proposed in \cite{tran2017bayesian}, and was evaluated on standard label-rich image classification datasets.
The authors of \cite{hauberg2016dreaming} learned class-conditional distributions by a diffeomorphism assumption. A concurrent preprint \cite{antoniou2017data} explored a Data Augmentation Generative Adversarial Network (DAGAN): the authors developed a completely different GAN model from the proposed DADA, whose generator does not depend on the classes and the discriminator is a vanilla real/fake one. Hence, it enables DAGAN to be applicable to unseen new classes for few-shot learning scenarios, different from our goal of improving fully-supervised classification. As we will see in experiments, deriving a stronger discriminator is critical in our target task. 


We also noticed an interesting benchmark study conducted in \cite{perez2017effectiveness} to compare among various data augmentation techniques, including very sophisticated generative models such as CycleGAN \cite{zhu2017unpaired}. Somewhat surprisingly, they found traditional ad-hoc augmentation techniques to be still able to outperform existing learning-based choices. Overall, enhancing small sample classification via learning-based data augmentation remains as an open and under-investigated problem. 

\vspace{0.2em}
\noindent\textit{Domain-specific Data Synthesis.} A number of works \cite{wang2015deepfont,le2017using,sixt2016rendergan,shrivastava2017learning,wangadversarial,jaderberg2014synthetic} explored the ``free'' generation of labeled synthetics examples to assist training. However, they either relied on extra information, e.g., 3D models of the subject, or were tailored for one special object class such as face or license plates. The synthesis could also be viewed as a special type of data augmentation that hinges on stronger forms of priori invariance knowledge. 
\vspace{0.2em}

\noindent\textit{Training Regularization.} A final option to fight against small datasets is to exploit variance reduction techniques for network design and training. Examples include dropout \cite{krizhevsky2012imagenet}, dropconnect \cite{wan2013regularization}, and enforcing compact structures on weights or connection patterns (e.g., sparsity) \cite{han2015learning}. Those techniques are for the general purposes of alleviating  overfitting, and they alone are unlikely to resolve the challenge of extremely low data regimes. 

\section{Technical Approach}

\subsection{Problem Formulation and Solution Overview}
Consider a general $k$-class classification problem. Suppose that we have a training set $\mathcal{D}=\{(x^{1},y^{1}),(x^{2},y^{2}),...,(x^{|\mathcal{D}|},y^{|\mathcal{D}|})\}$, where $x^{i}$ denotes a sample and $y^{i}$ a corresponding label, $ y^{i}\in \{1,2,...,k\}$. Our task is to learn a good \textbf{classifier} $C$ to predict the label $\hat{y}^{i} = C(x^{i})$, by minimizing the empirical risk objective $\frac{1}{|\mathcal{D}|}\sum^{|\mathcal{D}|}_{i=1} L(y^{i},\hat{y}^{i})$ over $\mathcal{D}$, $L$ being some loss function such as K-L divergence. As our goal, a good $C$ should generalize well on an unseen test set $\mathcal{T}$. In classical deep learning-based classification settings, $\mathcal{D}$ is large enough to ensure that goal. However in our extremely low data regimes, $|\mathcal{D}|$ can be too small to support robust learning of any complicated decision boundary, causing severe overfitting.

Data augmentation approaches seek an \textbf{augmenter} $A$, to synthesize a new set $\mathcal{D}'$ of augmented labeled data $(\bar{x}^{i},y^{i})$ from $(x^{i},y^{i})$, constituting the new augmented training set of size $|\mathcal{D}| + |\mathcal{D}'|$. Traditional choices of $A$, being mostly ad-hoc minor perturbations, are usually class-independent, i.e., constructing a sample-wise mapping from $x^{i}$ to $\bar{x}^{i}$ without taking into account the class distribution. Such mappings are usually limited to a small number of priori known, hand-crafted perturbations. They are not learned from data, and are not optimized towards finding classification boundaries. To further improve $A$, one may consider the inter-sample relationships \cite{hauberg2016dreaming}, as well as inter-class relationships in $\mathcal{D}$, where training a generative model $A$ over $(x^{i},y^{i})$ becomes a viable option.\\

\begin{figure}
\centering
\includegraphics[width=0.6\textwidth]{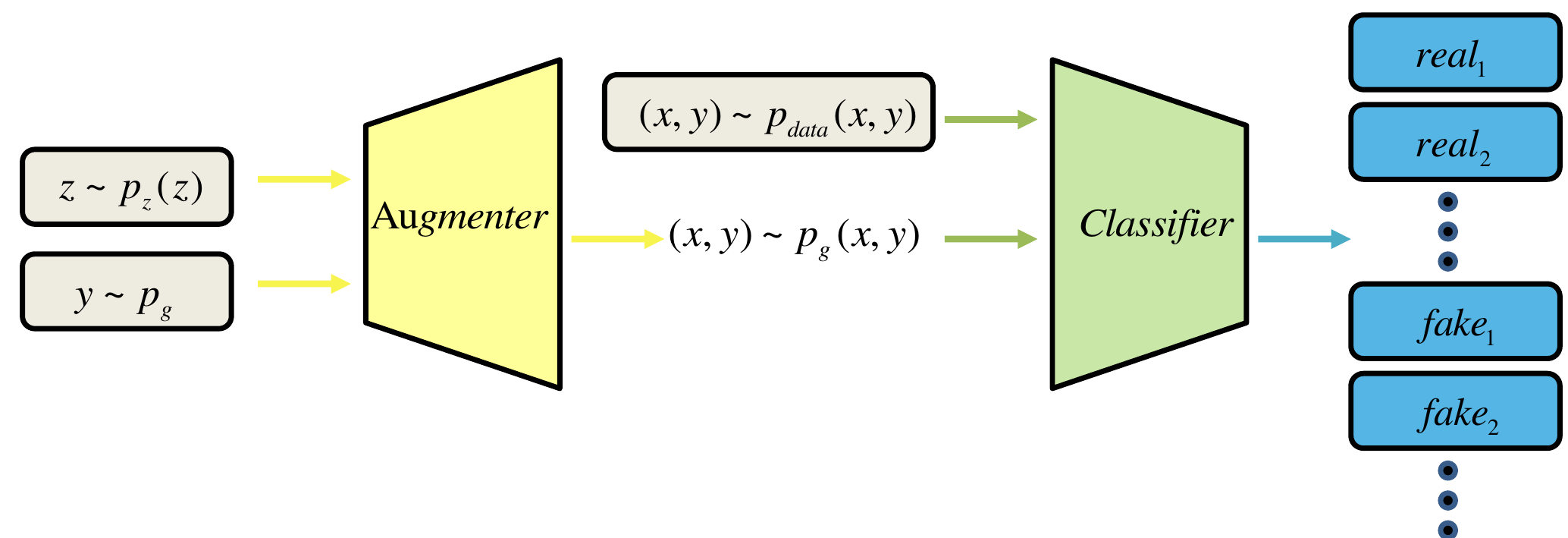}
\caption{\label{DADA}{An illustration of DADA.}}
\end{figure}
\vspace{-1em}
The conceptual framework of DADA is depicted in Figure \ref{DADA}. If taking a GAN point of view towards this, $A$ naturally resembles a generator: its inputs can be latent variables $z^{i}$ conditioned on $y^{i}$, and outputs $\bar{x}^{i}$ belonging to the same class $y^{i}$ but being \textbf{sufficiently diverse} from $x^{i}$. $C$ can act as the discriminator, if it will incorporate typical GAN's real-fake classification in addition to the target $k$-class classification. Ideally, the classifier $C$ should: (1) \textbf{be able to} correctly classify both real samples $x^{i}$ and augmented samples $\bar{x}^{i}$ into the correct class $y^{i}$; (2) \textbf{be unable to} distinguish $x^{i}$ and $\bar{x}^{i}$. The entire DADA framework of $A$ and $C$ can be jointly trained on $(x^{i},y^{i})$, whose procedure will bear similarities to training a class-conditional GAN. However, existing GANs may not fit the task well, due to the often low diversity of generated samples. We are hence motivated to introduce a novel loss function towards generating more diverse and class-specific samples.

\subsection{Going More Discriminative: From $k+1$ Loss to $2k$ Loss}
The discriminator of a vanilla, unsupervised GAN \cite{goodfellow2014generative} has only one output to indicate the probability of its input being a real sample. In \cite{salimans2016improved,odena2016semi}, the discriminator is extended with a semi-supervised fashion $k+1$ loss, whose output is a ($k+1$)-dimensional probabilistic vector: the first $k$ elements denote the probabilities of the input coming from the class 1, 2, ..., $k$ of real data; the ($k+1$)-th denotes its probability of belonging to the generated fake data. In that way, the generator simply has the semi-supervised classifier learned on additional unlabeled examples and supplied as a new ``generated'' class. In contrast, when in extremely low data regimes, we tend to be more ``economical'' on consuming data. 
We recognize that the unlabeled data provides weaker guidance than labeled data to learn the classification decision boundary. Therefore, if there is no real unlabeled data available and we can only generate from given limited labeled data, generating labeled data (if with quality) should benefit classifier learning more, compared to generating the same amount of unlabeled data. Further, the generated labeled samples should join force with the real labeled samples, and their decisions on the classification boundary should be well aligned. Motivated by the above design philosophy, we build a new $2k$ loss function, whose first group of $k$ outputs represent the probabilities of the input data from the class 1, 2, ..., $k$ of real data; its second group of $k$ outputs represent the probabilities of the input data from the class 1, 2, ..., $k$ of fake data. 

Since we use a class-conditional augmenter (generator), the label used to synthesize the augmented (fake) sample could be viewed to supply the ``ground truth'' for the second group of $k$ outputs. For example, for $k$ = 2, if the input datum is real and belongs to class 1, then its ground truth label is $[1, 0, 0, 0]$; otherwise if the input data is augmented conditionally on label of class 1, then its ground truth label is $[0, 0, 1, 0]$. During training, the K-L divergence is computed between the $2k$-length output and its ground truth label. For testing, we add the $i$-th and $(k+i)$-th elements of the $2k$ output to denote the probability of the input belonging to class $i$, $i = 1, 2, ..., k$. A comparison among loss functions for GANs including DADA is listed in Table \ref{loss}.

\vspace{-0.5em}

\begin{table}[htp!]
\centering
\caption{The comparison of loss functions among GAN discriminators}. 
\small
\begin{tabular}{c|c|c|c}
    \hline
    Model    & Class Number & Classes & Training Data \\
    \hline
    Vanilla GAN \cite{goodfellow2014generative} & 2 & real, fake & unlabeled only \\
    \hline
    Improved GAN \cite{salimans2016improved} & $k+1$ & $real_1$, ..., $real_k$; $fake$ & labeled + unlabeled\\
    \hline
    Proposed & $2k$ & $real_1$, ..., $real_k$; $fake_1$, ..., $fake_k$ & labeled only \\
    \hline
  \end{tabular}
\label{loss}
\end{table}

The detailed training algorithm for DADA is outlined in \textbf{supplementary}.

\vspace{-0.5em}

\section{Simulations}
\vspace{-1em}

To evaluate our approach, we first conduct a series of simulations on three widely adopted image classification benchmarks: CIFAR-10, CIFAR-100, and SVHN. We intentionally sample the given training data to simulate the extremely low data regimes, and compare the following training options. 1) C: directly train a classifier using the limited training data; 2) C\_augmented: perform traditional data augmentation (including rotation, translation and flipping), and then train a classifier; 3) DADA: the proposed data augmentation; 4) DADA\_augmented: first apply the same traditional augmentation as C\_augmented on the real samples, then perform DADA. We use absolutely \textbf{no unlabeled data or any pre-trained initialization} in training, different from the setting of most previous works. 
We use the original full test sets for evaluation. 
The network architectures that we used have been exhaustively tuned to ensure the best possible performance of all baselines in those unusually small training sets. 
Detailed configurations and hyperparameters, as well as visualized examples of augmented samples, are given in the \textbf{supplementary}.

\vspace{-1em}
\paragraph{CIFAR-10 and CIFAR-100}
The CIFAR-10 dataset consists of 60,000 color images at the resolution of 32$\times$32 in $k$ = 10 classes, with 5,000 images per class for training and 1,000 for testing. We sample the training data so that the amount of training images varies from 50 to 1,000 per class.

To illustrate the advantage of our proposed $2k$ loss, we also use the vanilla GAN \cite{goodfellow2014generative} (which adopt the $2$-class loss), as well as the Improved GAN \cite{salimans2016improved} (which adopt the $(k+1)$-class loss), as two additional baselines to augment samples. For the vanilla GAN, we train a separate generator \textit{for each class}. For Improved GAN, we provide only the labeled training data without using any unlabeled data: a different and more challenging setting than evaluated in \cite{salimans2016improved}. They work with traditional data augmentation too, similarly to the DADA\_augmented pipeline. For all compared methods, we generate samples so that the augmented dataset has 10 times the size of the given real labeled dataset. 

\begin{wrapfigure}{r}{0.50\textwidth}
\vspace{-2em}
\centering {
\includegraphics[width=200pt]{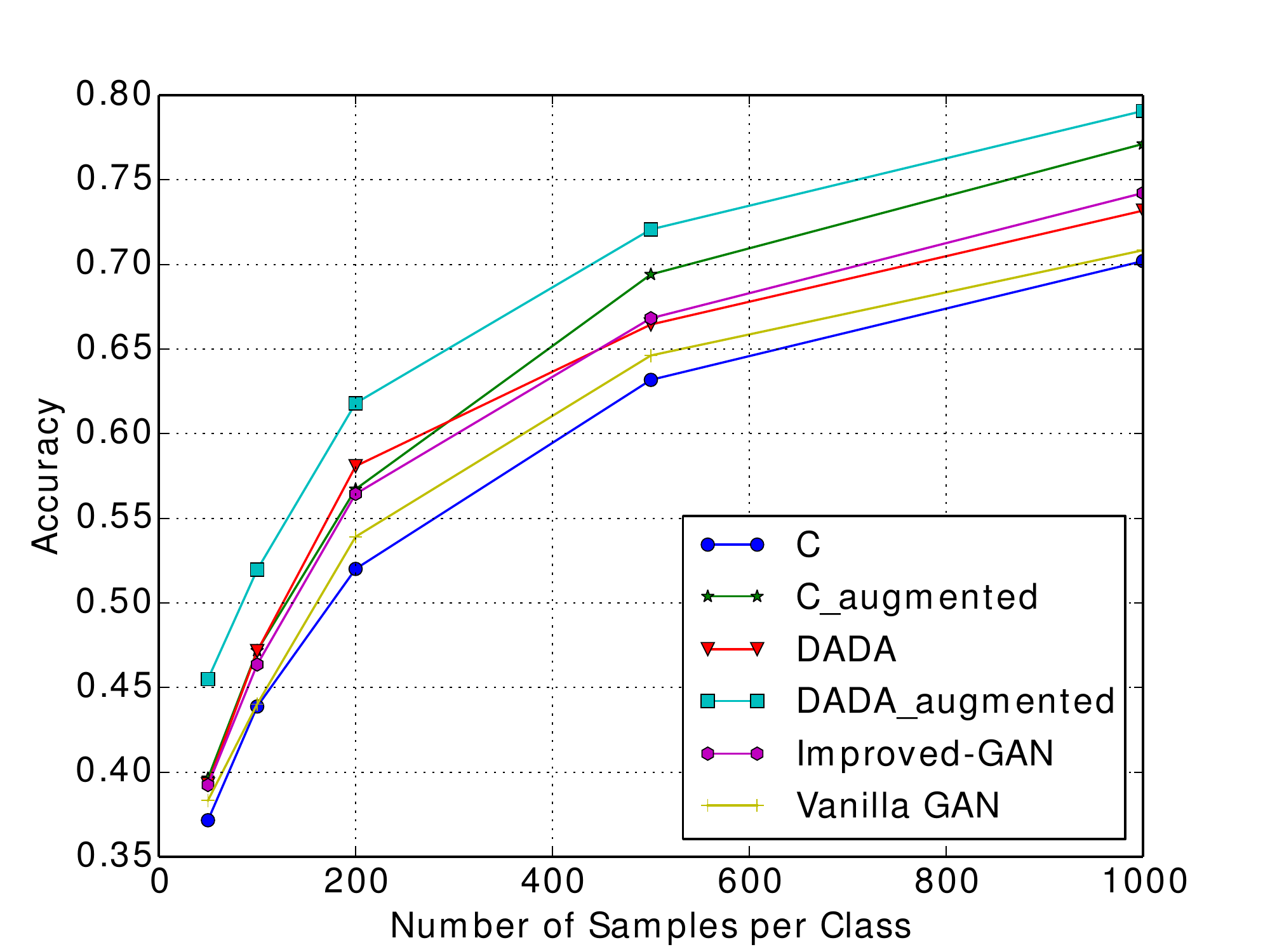}
}
\vspace{-1em}
\caption{Results on CIFAR-10, the test accuracy in different training settings with respect to the number of training images per class.} 
\vspace{-2em}
\label{cifar10_curve}
\end{wrapfigure}
Figure \ref{cifar10_curve} summarizes the performance of the compared methods. The vanilla GAN augmentation performs slightly better than the no-augmentation baseline, but the worst in all other data augmentation settings. It concurs with \cite{santurkar2017classification} that, though GAN can generate visually pleasing images, it does not naturally come with increased data diversity from a classification viewpoint. While improved GAN achieves superior performance, DADA (without using traditional augmentation) is able to outperform it at the smaller end of sample numbers (less than 400 per class). Comparing with vanilla GAN, Improved GAN and DADA\_augmented reveal that as the discriminator loss goes ``more discriminative'', the data augmentation becomes more effective along the way.  

Furthermore, DADA\_augmented is the best performer among all, and consistently surpass all other methods for the full range of [50, 1000] samples per class. It leads to around 8 percent top-1 accuracy improvement in the 500 labeled sample, 10 class subset, without relying on any unlabeled data. It also raises the top-1 performance to nearly 80\%, using only 10\% of the original training set (i.e. 1000 samples per class), again with neither pre-training nor unlabeled data.

It is worth pointing out that the traditional data augmentation C\_augmented presents a very competitive baseline here: it is next to DADA\_augmented, and becomes slightly inferior to DADA when the labeled samples are less than 300 per class, but is constantly better than all others. Further, integrating traditional data augmentation contributes to the consistent performance boost from DADA to DADA\_augmented. That testifies the value of empirical domain knowledge of invariance: they help considerably even learning-based augmentation is in place.

Finally, the comparison experiment is repeated on CIFAR-100. The results (see \textbf{supplementary}) are consistent with CIFAR-10, where DADA\_augmented achieves the best results and outperforms traditional data augmentation for at least 6\%, for all sample sizes. We also study the effects of DADA and traditional augmentation for deeper classifiers, such as ResNet-56 (see \textbf{supplementary}).\\



\setlength{\tabcolsep}{4pt}
\begin{table}
\begin{center}
\caption{Results on SVHN, the test accuracy in different training settings}
\label{svhn}
\begin{tabular}{llllll}
\hline\noalign{\smallskip}
\# Samples per class & $n=50$ & $n=80$ & $n=100$& $n=200$ & $n=500$\\
\noalign{\smallskip}
\hline
\noalign{\smallskip}
C  & 0.3767 & 0.6273 & 0.6867 & 0.8270 & \textbf{0.8974}\\
DADA & \textbf{0.4973} & \textbf{0.6818} & \textbf{0.7583} & \textbf{0.8472} & 0.8925\\
Improved-GAN & 0.4750 & 0.6729 & 0.7453 & 0.8364 & 0.8890\\
\hline
\end{tabular}
\vspace{-1.5em}
\end{center}
\end{table}

\setlength{\tabcolsep}{1.4pt}

\vspace{-2em}
\paragraph{SVHN}
SVHN is a digit recognition dataset, whose major challenge lies in that many images contain ``outlier'' digits but only the central digit is regarded as the target of recognition. As such, traditional data augmentation approaches such as translation or flipping may degrade training, and thus are excluded in this experiment. Table \ref{svhn} summarizes the results of using the proposed DADA (without combining traditional augmentation) in comparison with Improved GAN and the naive baseline of no data augmentation. It can be observed that, at extremely low data regimes, DADA again performs the best among the three. However, when a relatively large number of labeled samples are available (500 per class), DADA witnesses a slight negative impact on the accuracy compared to the naive baseline, but is still better than Improved GAN. We conjecture that this failure case is attributed to the ``outlier'' digits occurring frequently in SVNH that might hamper class-conditional generative modeling. We plan to explore more robust generators as future work to alleviate this problem.

We notice the larger margin of DADA (without augmentation) over Improved GAN on SVNH, compared to CIFAR-10. We conjecture the reason to be that SVNH has complicated perturbations (e.g., distracting digits), while CIFAR-10 is much ``cleaner'' in that sense (objects always lie in central foregrounds without other distractions). Thus on SVNH, the class information used by DADA could become more important in supervising the generation of high quality augmented samples, without being affected by perturbations.



\vspace{-1em}
\section{Experiments with Real-World Small Data}\label{sec_exp}
\vspace{-0.5em}

In this section, we discuss three real-data experiments which fall into extremely low data regimes. The data, not just labels, are difficult to collect and subject to high variability. We show that in these cases, the effects of transfer learning are limited, and/or even no ad-hoc data augmentation approach might be available to alleviate the difficulty to train deep networks. In comparison, DADA can be easily plugged in and boost the classification performance in all experiments. 

\vspace{-0.5em}
\subsection{Emotion Recognition from Facial Expressions: Comparison with Transfer Learning}
\vspace{-0.5em}
\noindent \textit{Background and Challenge.} Recognizing facial expressions is a topic of growing interests in the field of human-computer interaction. Among several public datasets in this field, the Karolinska Directed Emotional Faces (KDEF) dataset \cite{calvo2008facial} is a challenging benchmark consisting of rich facial variations (e.g., orientations, ethnicity, age, and gender), as well as relatively uniform distribution of the emotion classes. It has a total of 4,900 facial images collected from 70 individuals, displaying seven different facial expressions (happiness, fear, anger, disgust, surprise, sadness, neutrality). For each individual, the same expression is displayed twice and captured from 5 different angles. We choose images from the straight upfront angle in the first-time display only, forming a subset of 490 images for a 7-class classification problem. That certainly places us in an extremely low data regime. 

\vspace{0.1em}

\noindent \textit{Results and Analyses.} We use a random 5:2 split for the training and testing sets and pre-process the images by cropping and resizing the face regions to the resolution at 224$\times$224. We choose a VGG-16 model \cite{simonyan2014very} pre-trained on ImageNet as a baseline, which is re-trained and then tested on KDEF. We do not perform any traditional data augmentation, since each image is taken in a strictly-controlled setting. The baseline could be viewed as a transfer learning solution with ImageNet as the source domain. We then treat the pre-trained VGG-16 model as our classifier in DADA, and append it with an augmenter network (whose configuration is detailed in the \textbf{supplementary}). While the pre-trained VGG baseline gives rise to an accuracy of 82.86\%, DADA obtains a higher accuracy of 85.71\%. 

We also train vanilla GAN and Improved-GAN on this dataset, and have them  compare with DADA in the similar fair setting as in CIFAR-10. The vanilla GAN augmentation ends up with 83.27\% and Improved-GAN gets 84.03\%: both outperform transfer learning but stay inferior to DADA.

Transfer learning is often an effective choice for problems short of training data. But their effectiveness is limited when there are domain mismatches, even it is widely believed that ImageNet pre-trained models are highly transferable for most tasks. In this case, the gap between the source domain (ImageNet, general natural images) and the target domain (KDEF, facial images taken in lab environments) cannot be neglected. We advocate that learning-based data augmentation could boost the performance further on top of transfer learning, and their combination is more compelling. 

\vspace{-0.5em}
\subsection{Brain Signal Classification: No Domain Knowledge Can be Specified for Augmentation}
\vspace{-0.5em}

\noindent \textit{Background.} The classification of brain signals has found extensive applications in brain-computer interface, entertainment and rehabilitation engineering \cite{leeb2007brain}. Among various tasks, the electroencephalographic (EEG) signal classification problem has been widely explored. Existing approaches include band power method \cite{brodu2011comparative}, multivariate adaptive autoregressive (MVAAR) method \cite{anderson1998multivariate},  and independent component analysis (ICA) \cite{hung2005recognition}. Recent works \cite{ren2014convolutional,tabar2016novel} have explored CNNs in classifying EEG signals. However, the performance boost has been largely limited by the availability of labeled data. For example, the commonly used benchmark dataset 2b, a subset from the BCI Competition IV training set \cite{schlogl2003outcome}, includes only 400 trials. After several domain-specific pre-processing steps, each sample could be re-arranged into a $31 \times 32 \times 3$ image, where $3$ comes from the three EEG channels recorded (C3, Cz, and C4). They are collected from three sessions of motor imagery task experiments, and are to be classified into two classes of motions: right and left hand movements. We thus have a practical binary classification problem in extremely low data regimes.

\vspace{0.1em}

\noindent \textit{Challenge.} Unlike image classification problems discussed above, no straightforward knowledge-based, label-preserving augmentation has been proposed for EEG signals, nor has any data augmentation been applied in previous EEG classification works \cite{ren2014convolutional,tabar2016novel} to our best knowledge. Also, the noisy nature of brain signals discourages to manually add more perturbations. 
The major bottleneck for collecting EEG classification datasets lies in the expensive controlled data collection process itself, rather than the labeling (since subjects are required to perform designated motions in a monitored lab environment, the collected EEG signals are naturally labeled). Besides, the high variability of human subjects also limit the scope of transfer learning in EEG classification. The multi-fold challenges make EEG classification an appropriate user case and testbed for our proposed DADA approach. 

\vspace{0.1em}

\noindent \textit{Results and Analyses.} We follow \cite{tabar2016novel} to adopt the benchmark dataset 2b from BCI Competition IV training set \cite{schlogl2003outcome}. We train and test classification models, as well as DADA models separately for each of the nine subjects. We randomly select 90\% of 400 trials for training and the remaining 10\% for testing, and report the average accuracy of 10 runs. We treat each EEG input as a ``color image'' and adopt a mostly similar DADA model architecture as used for CIFAR-10 (except for changing class number)\footnote{Note that in the same channel of an EEG input, differently from a natural image, the signal coherence between vertical neighborhood (i.e., among different frequencies) is less than that between horizontal neighborhood (i.e., among different time stamps). The standard 2-D CNN is an oversimplified model here and could be improved by considering such anisotropy, which is the theme of our future work.}. We include three baselines reported in \cite{tabar2016novel} for comparison: directly classifying the inputs by SVM; a shallow CNN with one convolutional and one fully-connected layers (CNN); and a deeper CNN with one convolutional layer, concatenated with seven fully-connected layers pre-trained using stacked auto-encoder (CNN + SAE). Table \ref{eeg} shows the performance advantage of DADA over the competitive CNN-SAE method in all nine subjects, with an average accuracy margin of 1.7 percent.
 \begin{table}[tbp]
 \begin{center}
 \caption{The accuracy (\%) comparison on BCI Competition IV dataset 2b among SVM, CNN, CNN-SAE, and DADA, on subjects 1--9 and their average.}
 \label{eeg}
 \begin{tabular}{|c|c|c|c|c|c|c|c|c|c|c|c|}
 \hline
\textbf{Method} &  Sub. 1 &  Sub. 2  & Sub. 3  & Sub. 4  & Sub. 5 & Sub. 6 & Sub. 7 & Sub. 8 & Sub. 9 & \textbf{Average}\\
 \hline
 SVM & 71.8 & 64.5 & 69.3 & 93.0 & 77.5 & 72.5 & 68.0 & 69.8 & 65.0 & 72.4 \\
  \hline
 CNN & 74.5 & 64.3 & 71.8 & 94.5 & 79.5 & 75.0 & 70.5 & 71.8 & 71.0 & 74.8 \\
   \hline
 CNN-SAE & 76.0 & 65.8 & 75.3 & 95.3 & 83.0 & 79.5 & 74.5 & 75.3 & 75.3  & 77.6 \\
   \hline
 DADA & \textbf{76.6} & \textbf{66.8} & \textbf{75.6} & \textbf{96.5} & \textbf{83.2} & \textbf{80.2} & \textbf{77.0} & \textbf{78.6} & \textbf{79.6} & \textbf{79.3} \\
 \hline
 \end{tabular}
  \vspace{-2em}
 \end{center}
 \end{table}

\vspace{-0.5em}
\subsection{Tumor Classification: Comparison with Other Learning-based Augmentation}
In the existing learning-based data augmentation work \textit{Tanda} \cite{ratner2017learning}, most training comes with the help of unlabeled data. One exception we noticed is their experiment on the Curated Breast Imaging Subset of the Digital Database for Screening Mammography (CBIS-DDSM) \cite{clark2013cancer,heath2000digital,lee2016curated}, a medical image classification task whose data is expensive to collect besides labeling. Since both Tanda and DADA use the only available labeled dataset to learn data augmentation, we are able to perform a fair comparison on CBIS-DDSM between the two.
\vspace{1em}
\begin{wraptable}{r}{0.30\textwidth}
\center
\vspace{-1.5em}
\caption{Comparison between DADA and Tanda (in different training settings)}
\label{ddsm}
\begin{tabular}{ll}
\hline\noalign{\smallskip}
 Models & Acc \\
\noalign{\smallskip}
\hline
\noalign{\smallskip}
Tanda (MF)  & 0.5990 \\
Tanda (LSTM) & 0.6270 \\
DADA & 0.6196\\
DADA\_augmented  & \textbf{0.6549}\\
\hline
\end{tabular}
\setlength{\tabcolsep}{1.4pt}
\end{wraptable}
We follow the same configuration of the classifier used for CBIS-DDSM by Tanda: a four-layer all-convolution CNN with leaky ReLUs and batch normalization. We resize all medical images to 224 $\times$ 224. Note that Tanda heavily relies on hand-crafted augmentations: on DDMS, it uses many basic heuristics (crop, rotate, zoom, etc.) and several domain-specific transplantations. For DADA\_augmented, we apply only rotation, zooming, and contrast as the traditional augmentation pre-processing, to be consistent with the user-specified traditional augmentation modules in Tanda. We compare DADA and DADA\_augmented with two versions of Tanda using mean field (MF) and LSTM generators \cite{ratner2017learning}, with Table \ref{ddsm} showing the clear advantage of our approaches. 

\textit{What differentiates DADA and Tanda?} Tanda trains a generative sequence model over user-specified, knowledge-based transformation functions, while DADA is purely trained in a data-driven discriminative way. Unlike Tanda whose augmented samples always look like the naturalistic samples of each class, DADA may sometimes lead to augmented samples which are not visually close, but are optimized towards depicting the boundary between different classes. We display some ``un-naturalistic'' augmented samples found in the SVHN experiments in \textbf{supplementary}. Tanda also seems to benefit from the unlabeled data used in training, which ensures the transformed data points to be within the data distribution, while DADA can work robustly without unlabeled data (such as CBIS-DDSMF).

\vspace{-0.5em}
\section{Conclusion}
\vspace{-0.5em}
We present DADA, a learning-based data augmentation solution for training deep classifiers in extremely low data regimes. We leverage the power of GAN to generate new training data that both bear class labels and enhance diversity. A new $2k$ loss is elaborated for DADA and verified to boost the performance. We perform extensive simulations as well as three real-data experiments, where results all endorse the practical advantage of DADA. 
We anticipate that DADA can be applied into many real-world tasks, including satellite, military, and biomedical image/data classification.


\begin{thebibliography}{10}

\bibitem{krizhevsky2012imagenet}
Alex Krizhevsky, Ilya Sutskever, and Geoffrey~E Hinton.
\newblock Imagenet classification with deep convolutional neural networks.
\newblock In {\em NIPS}, pages 1097--1105, 2012.

\bibitem{perez2017effectiveness}
Luis Perez and Jason Wang.
\newblock The effectiveness of data augmentation in image classification using
  deep learning.
\newblock {\em arXiv preprint arXiv:1712.04621}, 2017.

\bibitem{han2015learning}
Song Han, Jeff Pool, John Tran, and William Dally.
\newblock Learning both weights and connections for efficient neural network.
\newblock In {\em NIPS}, pages 1135--1143, 2015.

\bibitem{erhan2010does}
Dumitru Erhan, Yoshua Bengio, Aaron Courville, Pierre-Antoine Manzagol, Pascal
  Vincent, and Samy Bengio.
\newblock Why does unsupervised pre-training help deep learning?
\newblock {\em Journal of Machine Learning Research}, 11(Feb):625--660, 2010.

\bibitem{raina2007self}
Rajat Raina, Alexis Battle, Honglak Lee, Benjamin Packer, and Andrew~Y Ng.
\newblock Self-taught learning: Transfer learning from unlabeled data.
\newblock In {\em ICML}, pages 759--766, 2007.

\bibitem{kingma2014semi}
Diederik~P Kingma, Shakir Mohamed, Danilo~Jimenez Rezende, and Max Welling.
\newblock Semi-supervised learning with deep generative models.
\newblock In {\em NIPS}, pages 3581--3589, 2014.

\bibitem{goodfellow2014generative}
Ian Goodfellow, Jean Pouget-Abadie, Mehdi Mirza, Bing Xu, David Warde-Farley,
  Sherjil Ozair, Aaron Courville, and Yoshua Bengio.
\newblock Generative adversarial nets.
\newblock In {\em NIPS}, pages 2672--2680, 2014.

\bibitem{mirza2014conditional}
Mehdi Mirza and Simon Osindero.
\newblock Conditional generative adversarial nets.
\newblock {\em arXiv preprint arXiv:1411.1784}, 2014.

\bibitem{odena2016semi}
Augustus Odena.
\newblock Semi-supervised learning with generative adversarial networks.
\newblock {\em arXiv preprint arXiv:1606.01583}, 2016.

\bibitem{salimans2016improved}
Tim Salimans, Ian Goodfellow, Wojciech Zaremba, Vicki Cheung, Alec Radford, and
  Xi~Chen.
\newblock Improved techniques for training {GANs}.
\newblock In {\em NIPS}, pages 2234--2242, 2016.

\bibitem{dai2017good}
Zihang Dai, Zhilin Yang, Fan Yang, William~W Cohen, and Ruslan~R Salakhutdinov.
\newblock Good semi-supervised learning that requires a bad {GAN}.
\newblock In {\em NIPS}, pages 6513--6523, 2017.

\bibitem{radford2015unsupervised}
Alec Radford, Luke Metz, and Soumith Chintala.
\newblock Unsupervised representation learning with deep convolutional
  generative adversarial networks.
\newblock {\em arXiv preprint arXiv:1511.06434}, 2015.

\bibitem{arora2017gans}
Sanjeev Arora and Yi~Zhang.
\newblock Do {GANs} actually learn the distribution? an empirical study.
\newblock {\em arXiv preprint arXiv:1706.08224}, 2017.

\bibitem{santurkar2017classification}
Shibani Santurkar, Ludwig Schmidt, and Aleksander Madry.
\newblock A classification-based perspective on {GAN} distributions.
\newblock {\em arXiv preprint arXiv:1711.00970}, 2017.

\bibitem{lin2015bilinear}
Tsung-Yu Lin, Aruni RoyChowdhury, and Subhransu Maji.
\newblock Bilinear {CNN} models for fine-grained visual recognition.
\newblock In {\em ICCV}, pages 1449--1457, 2015.

\bibitem{mehrotra2017generative}
Akshay Mehrotra and Ambedkar Dukkipati.
\newblock Generative adversarial residual pairwise networks for one shot
  learning.
\newblock {\em arXiv preprint arXiv:1703.08033}, 2017.

\bibitem{shu2017lifelong}
Lei Shu, Hu~Xu, and Bing Liu.
\newblock Lifelong learning crf for supervised aspect extraction.
\newblock {\em arXiv preprint arXiv:1705.00251}, 2017.

\bibitem{liu2017deep}
Bo~Liu, Ying Wei, Yu~Zhang, and Qiang Yang.
\newblock Deep neural networks for high dimension, low sample size data.
\newblock In {\em IJCAI}, pages 2287--2293, 2017.

\bibitem{vincent2010stacked}
Pascal Vincent, Hugo Larochelle, Isabelle Lajoie, Yoshua Bengio, and
  Pierre-Antoine Manzagol.
\newblock Stacked denoising autoencoders: Learning useful representations in a
  deep network with a local denoising criterion.
\newblock {\em Journal of Machine Learning Research}, 11(Dec):3371--3408, 2010.

\bibitem{lee2013pseudo}
Dong-Hyun Lee.
\newblock Pseudo-label: The simple and efficient semi-supervised learning
  method for deep neural networks.
\newblock In {\em Workshop on Challenges in Representation Learning, ICML},
  volume~3, page~2, 2013.

\bibitem{rasmus2015semi}
Antti Rasmus, Mathias Berglund, Mikko Honkala, Harri Valpola, and Tapani Raiko.
\newblock Semi-supervised learning with ladder networks.
\newblock In {\em NIPS}, pages 3546--3554, 2015.

\bibitem{wagner2013learning}
Raimar Wagner, Markus Thom, Roland Schweiger, Gunther Palm, and Albrecht
  Rothermel.
\newblock Learning convolutional neural networks from few samples.
\newblock In {\em IJCNN}, pages 1--7, 2013.

\bibitem{ratner2017learning}
Alexander~J Ratner, Henry Ehrenberg, Zeshan Hussain, Jared Dunnmon, and
  Christopher R{\'e}.
\newblock Learning to compose domain-specific transformations for data
  augmentation.
\newblock In {\em Advances in Neural Information Processing Systems}, pages
  3239--3249, 2017.

\bibitem{tran2017bayesian}
Toan Tran, Trung Pham, Gustavo Carneiro, Lyle Palmer, and Ian Reid.
\newblock A bayesian data augmentation approach for learning deep models.
\newblock In {\em Advances in Neural Information Processing Systems}, pages
  2797--2806, 2017.

\bibitem{hauberg2016dreaming}
S{\o}ren Hauberg, Oren Freifeld, Anders Boesen~Lindbo Larsen, John Fisher, and
  Lars Hansen.
\newblock Dreaming more data: Class-dependent distributions over
  diffeomorphisms for learned data augmentation.
\newblock In {\em Artificial Intelligence and Statistics}, pages 342--350,
  2016.

\bibitem{antoniou2017data}
Antreas Antoniou, Amos Storkey, and Harrison Edwards.
\newblock Data augmentation generative adversarial networks.
\newblock {\em arXiv preprint arXiv:1711.04340}, 2017.

\bibitem{zhu2017unpaired}
Jun-Yan Zhu, Taesung Park, Phillip Isola, and Alexei~A Efros.
\newblock Unpaired image-to-image translation using cycle-consistent
  adversarial networks.
\newblock {\em arXiv preprint arXiv:1703.10593}, 2017.

\bibitem{wang2015deepfont}
Zhangyang Wang, Jianchao Yang, Hailin Jin, Eli Shechtman, Aseem Agarwala,
  Jonathan Brandt, and Thomas~S Huang.
\newblock Deepfont: Identify your font from an image.
\newblock In {\em Proceedings of the 23rd ACM international conference on
  Multimedia}, pages 451--459. ACM, 2015.

\bibitem{le2017using}
Tuan~Anh Le, Atilim~Giine{\c{s}} Baydin, Robert Zinkov, and Frank Wood.
\newblock Using synthetic data to train neural networks is model-based
  reasoning.
\newblock In {\em IJCNN}, pages 3514--3521, 2017.

\bibitem{sixt2016rendergan}
Leon Sixt, Benjamin Wild, and Tim Landgraf.
\newblock Rendergan: Generating realistic labeled data.
\newblock {\em arXiv preprint arXiv:1611.01331}, 2016.

\bibitem{shrivastava2017learning}
Ashish Shrivastava, Tomas Pfister, Oncel Tuzel, Josh Susskind, Wenda Wang, and
  Russ Webb.
\newblock Learning from simulated and unsupervised images through adversarial
  training.
\newblock In {\em CVPR}, volume~3, page~6, 2017.

\bibitem{wangadversarial}
Xinlong Wang, Zhipeng Man, Mingyu You, and Chunhua Shen.
\newblock Adversarial generation of training examples: Applications to moving
  vehicle license plate recognition.
\newblock {\em arXiv preprint arXiv:1707.03124}, 2017.

\bibitem{jaderberg2014synthetic}
Max Jaderberg, Karen Simonyan, Andrea Vedaldi, and Andrew Zisserman.
\newblock Synthetic data and artificial neural networks for natural scene text
  recognition.
\newblock {\em arXiv preprint arXiv:1406.2227}, 2014.

\bibitem{wan2013regularization}
Li~Wan, Matthew Zeiler, Sixin Zhang, Yann Le~Cun, and Rob Fergus.
\newblock Regularization of neural networks using dropconnect.
\newblock In {\em ICML}, pages 1058--1066, 2013.

\bibitem{calvo2008facial}
Manuel~G Calvo and Daniel Lundqvist.
\newblock Facial expressions of emotion ({KDEF}): Identification under
  different display-duration conditions.
\newblock {\em Behavior Research Methods}, 40(1):109--115, 2008.

\bibitem{simonyan2014very}
Karen Simonyan and Andrew Zisserman.
\newblock Very deep convolutional networks for large-scale image recognition.
\newblock {\em arXiv preprint arXiv:1409.1556}, 2014.

\bibitem{leeb2007brain}
Robert Leeb, Felix Lee, Claudia Keinrath, Reinhold Scherer, Horst Bischof, and
  Gert Pfurtscheller.
\newblock Brain--computer communication: Motivation, aim, and impact of
  exploring a virtual apartment.
\newblock {\em IEEE Transactions on Neural Systems and Rehabilitation
  Engineering}, 15(4):473--482, 2007.

\bibitem{brodu2011comparative}
Nicolas Brodu, Fabien Lotte, and Anatole L{\'e}cuyer.
\newblock Comparative study of band-power extraction techniques for motor
  imagery classification.
\newblock In {\em IEEE Symposium on Computational Intelligence, Cognitive
  Algorithms, Mind, and Brain (CCMB)}, 2011.

\bibitem{anderson1998multivariate}
Charles~W Anderson, Erik~A Stolz, and Sanyogita Shamsunder.
\newblock Multivariate autoregressive models for classification of spontaneous
  electroencephalographic signals during mental tasks.
\newblock {\em IEEE Transactions on Biomedical Engineering}, 45(3):277--286,
  1998.

\bibitem{hung2005recognition}
Chih-I Hung, Po-Lei Lee, Yu-Te Wu, Li-Fen Chen, Tzu-Chen Yeh, and Jen-Chuen
  Hsieh.
\newblock Recognition of motor imagery electroencephalography using independent
  component analysis and machine classifiers.
\newblock {\em Annals of Biomedical Engineering}, 33(8):1053--1070, 2005.

\bibitem{ren2014convolutional}
Yuanfang Ren and Yan Wu.
\newblock Convolutional deep belief networks for feature extraction of {EEG}
  signal.
\newblock In {\em IJCNN}, 2014.

\bibitem{tabar2016novel}
Yousef~Rezaei Tabar and Ugur Halici.
\newblock A novel deep learning approach for classification of {EEG} motor
  imagery signals.
\newblock {\em Journal of Neural Engineering}, 14(1):016003, 2016.

\bibitem{schlogl2003outcome}
Alois Schl{\"o}gl.
\newblock Outcome of the {BCI}-competition 2003 on the {Graz} data set.
\newblock Technical report, Graz University of Technology, 2003.

\bibitem{clark2013cancer}
Kenneth Clark, Bruce Vendt, Kirk Smith, John Freymann, Justin Kirby, Paul
  Koppel, Stephen Moore, Stanley Phillips, David Maffitt, Michael Pringle,
  et~al.
\newblock The cancer imaging archive (tcia): maintaining and operating a public
  information repository.
\newblock {\em Journal of digital imaging}, 26(6):1045--1057, 2013.

\bibitem{heath2000digital}
M~Heath, K~Bowyer, D~Kopans, R~Moore, and P~Kegelmeyer.
\newblock The digital database for screening mammography.
\newblock {\em Digital mammography}, pages 431--434, 2000.

\bibitem{lee2016curated}
R~Sawyer Lee, F~Gimenez, A~Hoogi, and D~Rubin.
\newblock Curated breast imaging subset of ddsm.
\newblock {\em The Cancer Imaging Archive}, 2016.

\end{thebibliography}

\appendix
\section{Training Algorithm Details}
Due to the different roles that the classifier have to play, the training procedure of DADA is divided into two different phases. In training phase I, which we call \emph{Generation training}, the classifier and the augmenter compete with each other like in the vanilla GAN. The difference is that their competition is conditioned on the specific class, rather than the whole data set. The augmenter attempts to generate realistic data to cheat the classifier within a specific class, while the classifier endeavors to distinguish the fake data from the real within a specific class. The game between the two players will have its optimum only if $ p_{data}(x|y)=p_{g}(x|y)$. Thus, the optimal classifier has $C(x|y)=p_{data}(x|y)/(p_{g}(x|y)+p_{data}(x|y))=1/2$, indicating that the augmenter is trained well enough so that the classifier can not discriminate them.  

\begin{algorithm}
\caption{Minibatch stochastic gradient descent training of DADA}\label{algorithm}
\begin{algorithmic}[1]
\REQUIRE The training epochs $\mathcal{K}_{G}$, $\mathcal{K}_{C}$ in phase \uppercase\expandafter{\romannumeral1} and phase \uppercase\expandafter{\romannumeral2}, the training set $\mathcal{D}$, the test set $\mathcal{T}$, the batch size $\mathcal{B}$
\FOR{number of epochs $\mathcal{K}_{G}$}
\STATE Sample a batch of pairs $(z,y)$, $z\sim p_{z}(z),y\sim p_{g}$, a batch of pairs $(x,y)\sim p_{data}(x,y)$.
\STATE Update the classifier by performing stochastic gradient descent on $L_{C}^{\uppercase\expandafter{\romannumeral1}}$
\FOR{number of epochs}
\STATE Sample a batch of pairs $(x,y)\sim p_{data}(x,y)$, a batch of pairs $(z,y)$, $z\sim p_{z}(z)$, keep $y$ the same with the true data 
\STATE Update the generator by performing stochastic gradient descent on $L^{I}_{G}$
\ENDFOR
\ENDFOR
\FOR{number of epochs $\mathcal{K}_{C}$}
\STATE Sample a batch of pairs $(z,y)$, $z\sim p_{z}(z),y\sim p_{g}$, a batch of pairs $(x,y)\sim p_{data}(x,y)$.
\STATE Update the classifier by performing stochastic gradient descent on $L_{C}^{\uppercase\expandafter{\romannumeral2}}$
\ENDFOR
\end{algorithmic}
\end{algorithm}
Similar to the vanilla GAN formulation, the loss functions of the augmenter and the classifier in training phase \uppercase\expandafter{\romannumeral1} are:
\begin{align}\label{1}
L_{C}^{\uppercase\expandafter{\romannumeral1}}=&-\mathbf{E}_{x,y\sim p_{data}(x,y)}\log[p(y|x,y<k+1)]\nonumber\\&-\mathbf{E}_{x,y\sim p_{g}(x,y)}\log [p(y|x,k<y<2k+1)]
\end{align}
\begin{equation}\label{2}
\mathcal{L}^{I}_{G}=-\mathbf{E}_{x,y\sim p_{g}(x,y)}\log [p(y-k|x,k<y<2k+1)]
\end{equation}
Based on the observation of the Improved-GAN that the feature matching technique can help improve the classification performance of the generated samples, we make some modifications on this training strategy. Because we desire not only the features of the training data and the generated data to be as similar as possible, but also the training data and the generated data to be within a specific class, the modified version of feature matching is formulated as:
\begin{equation}\label{3}
\mathcal{L}_{fm}=\|\mathbf{E}_{x,y\sim p_{data}(x,y)}f(x|y)-\mathbf{E}_{z\sim p_{z}(z),y\sim p_{c}}f(G(z,y)|y) \|
\end{equation}
Here $f(x)$ denotes activations on an intermediate layer of the classifier. We keep $p_c$ the same to the true data label. With the regularization of feature matching, the objective function of generator in training phase \uppercase\expandafter{\romannumeral1} is hence:
\begin{equation}\label{4}
L^{I}_{G}=\mathcal{L}^{I}_{G}+\lambda \mathcal{L}_{fm}
\end{equation}

Once the training phase \uppercase\expandafter{\romannumeral1} is finished, assuming that the generator can capture the class-wise data distribution, then it comes to the training phase \uppercase\expandafter{\romannumeral2} called \emph{Classification training}. In this phase, the generator is fixed just as a data provider. We only train the classifier on the generated data and the real training data. The loss function of the classifier in training phase \uppercase\expandafter{\romannumeral2} can be written as:
\begin{equation}\label{5}
L_{C}^{\uppercase\expandafter{\romannumeral2}}=\mathcal{L}_{data}+\mathcal{L}_{gen}
\end{equation}
where,
\begin{equation}\label{6}
\mathcal{L}_{data}=-\mathbf{E}_{x,y\sim p_{data}(x,y)}\log[p(y|x,y<k+1)+p(y+k|x,y<k+1)]
\end{equation}
and
\begin{equation}\label{7}
\mathcal{L}_{gen}=-\mathbf{E}_{x,y\sim p_{g}(x,y)}\log[p(y|x,k<y<2k+1)+p(y-k|x,k<y<2k+1)]
\end{equation}
The entire training procedure is summarized in Algorithm \ref{algorithm}, where the two training stages are described as two for-loops.

\section{Examples of Generated Samples}
\begin{figure}
\centering
\includegraphics[scale=0.7]{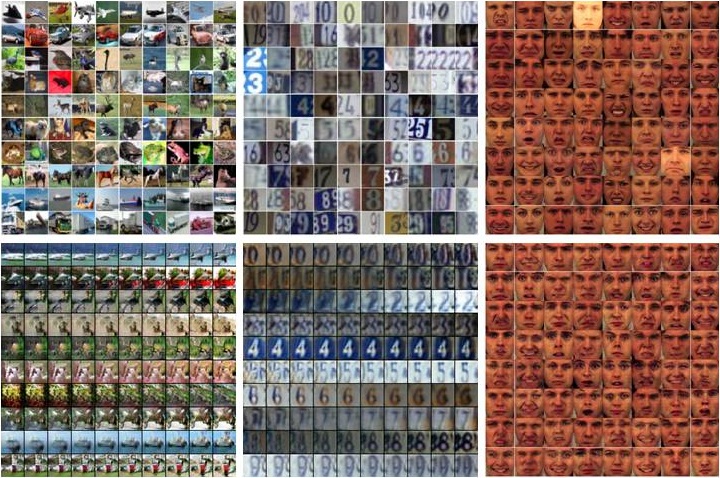}
\caption{Visual comparison between (top) the samples from training data and (bottom) the generated samples by DADA. From left to right: CIFAR-10, SVHN, and KDEF. For CIFAR-10 and SVHN, \textbf{each row corresponds to one class and within each row, the images are generated by linearly interpolated random vectors} (except the face image by random noise).}
\label{generation}
\end{figure}

For visual inspection, Figure \ref{generation} depicts a few samples of real images as well as the generated samples by DADA (by interpolating latent codes). In these figures, the CIFAR-10 and SVHN examples correspond to the extremely low data regimes (200 per class). Therefore, we do not anticipate the generated samples to contain large visual variety. Recall that our objective is to augment data for training a better classifier, the generated images can be observed to faithfully belong to the classes that they are conditioned on, and present certain variances, which fulfills our goal.

\begin{figure}
\centering
\includegraphics[scale=0.7]{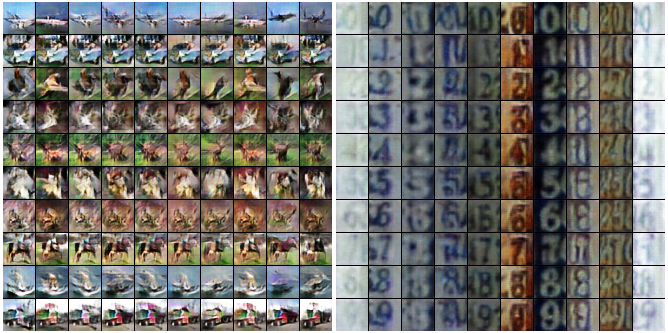}
\caption{Left: visual diversity for CIFAR-10; Right: ``un-naturalistic'' samples for SVHN. \textbf{Each row corresponds to one class and each column uses same latent code.}}
\label{generation1}
\end{figure}

Figure \ref{generation1} displays the diversity of generated samples as well as some ``un-naturalistic'' examples found in the SVHN experiments. For example, in the fourth row (class ``3''), several examples look like ``8''; also in the sixth row (class ``5''), one might find examples in part resembling ``6'' or ``8''. Those ``un-naturalistic'' examples are obviously close to boundaries between confusing digit classes. 

\section{CIFAR-10 Experiments with Deeper Classifiers}
How far we can go with DADA to train even deeper classifiers in extremely low-data regimes? To test that, we replace the default classifier in our CIFAR-10 experiment with a ResNet-56 model, and train it with DADA. Table \ref{cifar10} compares DADA\_augmented with traditional augmentation at $n$ = 500 and $1000$. DADA\_augmented remains superior, and all results show improvements over the original numbers in Figure 2 (main text). However, for $n$ less than 200, both methods show severe overfitting and their performance degrade fast. That reminds us that deeper classifers may not be the right choice when the data and label are both too limited. 

\section{CIFAR-100 Experiments}
CIFAR-100 is an extended and more difficult version of CIFAR-10, containing 600 images (500 for training and 100 for testing) per each of the 100 classes. Similar to the results of CIFAR-10, DADA boosts the classification accuracy, and at the small sample choices DADA (without augmentation) has a clear advantage over traditional augmentation. Their combination DADA\_augmented achieves the best results and outperforms the others significantly (more than 6 percent compared to only using traditional data augmentation); see in Table \ref{cifar100}.\\

\setlength{\tabcolsep}{4pt}
\begin{table}
\begin{center}
\caption{Results on CIFAR-10 with a ResNet-56 classifier.}
\label{cifar10}
\begin{tabular}{lll}
\hline\noalign{\smallskip}
\# Samples per class &$n=500$ & $n=1000$ \\
\noalign{\smallskip}
\hline
\noalign{\smallskip}
C\_augmented&0.7797 & 0.8247 \\
DADA\_augmented&\textbf{0.7927} & \textbf{0.8325}\\
\hline
\end{tabular}
\end{center}
\end{table}

\setlength{\tabcolsep}{1.4pt}

\setlength{\tabcolsep}{4pt}
\begin{table}
\begin{center}
\caption{Results on CIFAR-100, the test accuracy in different training settings.}
\label{cifar100}
\begin{tabular}{lllll}
\hline\noalign{\smallskip}
\# Samples per class &$n=50$ & $n=80$ & $n=100$ & $n=200$\\
\noalign{\smallskip}
\hline
\noalign{\smallskip}
C & 0.2991& 0.3742 & 0.4079 & 0.5189 \\
C\_augmented&0.3055 & 0.3806 & 0.4214 & 0.5426\\
DADA& 0.3718& 0.4337 & 0.4770 & 0.5630 \\
DADA\_augmented&\textbf{0.3727} & \textbf{0.4443} & \textbf{0.4856} & \textbf{0.5807}\\
\hline
\end{tabular}
\end{center}
\end{table}

\setlength{\tabcolsep}{1.4pt}
\section{Network architecture}
Here we give the detailed configurations and hyperparameters of our models. Table \ref{config} shows the details of CIFAR-10 and SVHN. The configurations of CIFAR-100 and EEG are almost the same as those of CIFAR-10 and SVHN except that the numbers of classes are different. The classifier in the KDEF experiment follows the VGG-16 architecture except that we append a weight normalization layer after each convolution layer. Table \ref{config_kdef} shows the details of KDEF generator (augmenter).

\vspace{2em}
\setlength{\tabcolsep}{4pt}
\begin{table}[!htbp]
\begin{center}
\caption{CIFAR-10 and SVHN network architectures of Augmenter and Classifier. The Vanilla Classifier is exactly the same as the Classifier. Here, A, C, T-Conv, Conv, NIN, BN, WN, and NL stand for Augmenter, Classifier, Transposed-Convolution, Convolution, Network in Network, Batch Normalization, Weight Normalization and Nonlinearity, respectively. We train our DADA model for totally 800 epochs: $\mathcal{K}_{G}$ = 200, $\mathcal{K}_{C}$ = 600 with Adam ($l = 0.0003$, $\beta_{1} = 0.9$, $\beta_{2} = 0.999$).}
\label{config}
\begin{tabular}{rllllll}
\hline\noalign{\smallskip}
Operation & Kernel & Strides & Feature maps & BN/WN? & Dropout & NL\\
\noalign{\smallskip}
\hline
\noalign{\smallskip}
A-110x1x1 input\\
Linear&N/A & N/A & 8192 & \checkmark & 0.0 & ReLU\\
Reshape&N/A & N/A & 512 & $\times$ & 0.0 & None\\
Concatenation&N/A & N/A  & 522& $\times$ & 0.0 & None \\
T-Conv & 5 x 5 & 2 x 2 & 256 & \checkmark & 0.0 & ReLU \\
Concatenation & N/A & N/A & 266 & $\times$ & 0.0& None\\
T-Conv & 5 x 5 & 2 x 2 & 128 & \checkmark & 0.0 & ReLU \\
Concatenation & N/A & N/A & 138 & $\times$ & 0.0 & None \\
T-Conv & 5 x 5 & 2 x 2 & 3 & \checkmark & 0.0 & Tanh \\
C-32x32x3 input \\
GaussianLayer & N/A & N/A & N/A & $\times$ & 0.0 & None\\
Conv & 3 x 3 & 1 x 1& 96 & \checkmark & 0.0 & LReLU \\
Conv & 3 x 3 & 1 x 1& 96 & \checkmark & 0.0 & LReLU \\
Conv & 3 x 3 & 2 x 2& 96 & \checkmark & 0.5 & LReLU \\
Conv & 3 x 3 & 1 x 1& 192 & \checkmark & 0.0 & LReLU \\
Conv & 3 x 3 & 1 x 1& 192 & \checkmark & 0.0 & LReLU \\
Conv & 3 x 3 & 2 x 2& 192 & \checkmark & 0.5 & LReLU \\
Conv & 3 x 3 & 1 x 1& 192 & \checkmark & 0.0 & LReLU \\
NIN & N/A & N/A & 192 & \checkmark & 0.0 & LReLU \\
NIN & N/A & N/A & 192 & \checkmark & 0.0 & LReLU \\
Global Pooling & N/A & N/A & 192 & $\times$ & 0.0 & None \\
Linear & N/A & N/A & 20 & \checkmark & 0.0 & None
\\
\hline
\end{tabular}
\end{center}
\end{table}
\setlength{\tabcolsep}{1.4pt}

\setlength{\tabcolsep}{4pt}
\begin{table}
\begin{center}
\caption{KDEF network architectures of Augmenter. The classifier is exactly the same as that of VGG-16 except that we append a weight normalization layer after each convolution operation. Here, A, C, T-Conv, Conv, NIN, BN, WN, and NL stand for Augmenter, Classifier, Transposed-Convolution, Convolution, Network in Network, Batch Normalization, Weight Normalization and Nonlinearity, respectively. We train our DADA model for totally 300 epochs: $\mathcal{K}_{G}$ = 100, $\mathcal{K}_{C}$ = 200 with Adam ($l = 0.0003$, $\beta_{1} = 0.9$, $\beta_{2} = 0.999$).}
\label{config_kdef}
\begin{tabular}{rllllll}
\hline\noalign{\smallskip}
Operation & Kernel & Strides & Feature maps & BN/WN? & Dropout & NL\\
\noalign{\smallskip}
\hline
\noalign{\smallskip}
A-107x1x1 input\\
Linear&N/A & N/A & 25088 & \checkmark & 0.0 & ReLU\\
Reshape&N/A & N/A & 512 & $\times$ & 0.0 & None\\
Concatenation&N/A & N/A  & 519& $\times$ & 0.0 & None \\
T-Conv & 5 x 5 & 2 x 2 & 512 & \checkmark & 0.0 & ReLU \\
Concatenation & N/A & N/A & 519 & $\times$ & 0.0& None\\
T-Conv & 5 x 5 & 2 x 2 & 256 & \checkmark & 0.0 & ReLU \\
Concatenation & N/A & N/A & 263 & $\times$ & 0.0 & None \\
T-Conv & 5 x 5 & 2 x 2 & 256 & \checkmark & 0.0 & ReLU \\
Concatenation & N/A & N/A & 263 & $\times$ & 0.0 & None \\
T-Conv & 5 x 5 & 2 x 2 & 128 & \checkmark & 0.0 & ReLU \\
Concatenation & N/A & N/A & 135 & $\times$ & 0.0 & None \\
T-Conv & 5 x 5 & 2 x 2 & 3 & \checkmark & 0.0 & Tanh \\
\hline
\end{tabular}
\end{center}
\end{table}
\setlength{\tabcolsep}{1.4pt}
\end{document}